\documentclass[10pt,twocolumn,letterpaper]{article}
\usepackage[margin=0.75in,columnsep=0.28in]{geometry}
\usepackage{newtxtext,newtxmath}
\usepackage[T1]{fontenc}
\usepackage{microtype,graphicx,booktabs,longtable,caption,subcaption,xcolor,enumitem,amsmath}
\usepackage[hidelinks]{hyperref}
\usepackage{titlesec,float}
\titleformat{\section}{\large\bfseries}{\thesection}{0.6em}{}
\titleformat{\subsection}{\normalsize\bfseries}{\thesubsection}{0.6em}{}
\setlist{nosep,leftmargin=1.2em}
\definecolor{oski}{HTML}{3F7D58}
\newcommand{\figw}{\columnwidth}
\newcommand{\sessiontable}{%
\begin{table*}[t]\centering\small
\caption{The five observed sessions. Face px: median face size. Camera: share of the recording with $\geq$3 faces in view. Recognised: students named for $\geq$3\,s; glimpsed: named for under 3\,s. Wrong sec.: people matched from another section (certain errors, any duration). Minutes of instructor and non-instructor speech from voiceprints. Calls: instructor name calls detected. Named: student comments attributed.}
\label{tab:sessions}
\resizebox{\textwidth}{!}{\begin{tabular}{lrrrrrrrrrr}\toprule
Class & Video & Face px & Camera & Recognised & Glimpsed & Wrong sec. & Instr. min & Student min & Calls & Named\\\midrule
Sep 22 & 1024$\times$576 & 29 & 33\% & 42 & 1 & 1 & 50 & 39 & 10 & 32 \\
Sep 24 & 1024$\times$576 & 30 & 26\% & 41 & 3 & 4 & 52 & 22 & 13 & 30 \\
Sep 29 & 1024$\times$576 & 33 & 22\% & 36 & 1 & 4 & 57 & 26 & 4 & 12 \\
Oct 6 & 1024$\times$576 & 33 & 24\% & 33 & 3 & 0 & 33 & 57 & 4 & 21 \\
Oct 8 & 1920$\times$1080 & 57 & 11\% & 41 & 2 & 1 & 37 & 11 & 3 & 7 \\
\bottomrule\end{tabular}}\end{table*}}

\newcommand{\idtable}{%
\begin{table}[t]\centering\small
\caption{Identification on the 1080p session (5,785 faces $\geq$24\,px, sampled every 2\,s) against a 229-student, four-section gallery. Certain errors are matches to another section; estimated correctness scales them by 229/172.}
\label{tab:id}
\resizebox{\columnwidth}{!}{\begin{tabular}{rrrrrrr}\toprule
$\tau$ & Named & Share & Certain err. & Est.\ correct & Students & Wrong people\\\midrule
0.40 & 2,579 & 44.6\% & 19 & 99.0\% & 44 & 16 \\
0.45 & 2,025 & 35.0\% & 3 & 99.8\% & 43 & 3 \\
0.50 & 1,393 & 24.1\% & 0 & 100.0\% & 42 & 0 \\
0.55 & 765 & 13.2\% & 0 & 100.0\% & 37 & 0 \\
0.60 & 304 & 5.3\% & 0 & 100.0\% & 27 & 0 \\
\bottomrule\end{tabular}}\end{table}}

\newcommand{\vertable}{%
\begin{table}[t]\centering\small
\caption{Verification from the classroom's own footage: 2,733 same-seat pairs (1\,s apart) and 19,801 same-frame pairs. The last column is the chance an unknown face clears the bar against at least one of 70 enrolled students.}
\label{tab:ver}
\begin{tabular}{rrrr}\toprule
$\tau$ & Same person & False match / pair & False match vs 70\\\midrule
0.300 & 99.5\% & 3.596\% & 92.3\% \\
0.363 & 99.2\% & 0.828\% & 44.1\% \\
0.400 & 98.8\% & 0.278\% & 17.7\% \\
0.450 & 97.9\% & 0.081\% & 5.5\% \\
0.500 & 96.9\% & 0.010\% & 0.7\% \\
\bottomrule\end{tabular}\end{table}}

\newcommand{\studenttable}{%
\begin{footnotesize}\setlength{\tabcolsep}{5pt}\renewcommand{\arraystretch}{0.95}
\begin{longtable}{lrrrrrrrrrl}
\caption{Every student, by anonymous id. Seconds recognised on camera per class (``--'' = not recognised), classes present ($\geq$3\,s), named comments ($^\dagger$ includes voice matches from a class the student was not seen in), named talk seconds, times called on, and whether a voiceprint was learned.}\label{tab:students}\\\toprule
Student & Sep 22 & Sep 24 & Sep 29 & Oct 6 & Oct 8 & Cls & Cmts & Talk s & Calls & Voice\\\midrule\endfirsthead
\toprule Student & Sep 22 & Sep 24 & Sep 29 & Oct 6 & Oct 8 & Cls & Cmts & Talk s & Calls & Voice\\\midrule\endhead
S01 & 134 & 25 & 55 & 65 & -- & 4 & 0 & 0 & 1 &  \\
S02 & 404 & 118 & 158 & 177 & 211 & 5 & 0 & 0 & 0 &  \\
S03 & -- & -- & -- & -- & 4 & 1 & 0 & 0 & 1 &  \\
S04 & 342 & 207 & 252 & 477 & -- & 4 & 0 & 0 & 0 &  \\
S05 & -- & -- & -- & -- & 7 & 1 & 9$^\dagger$ & 264 & 1 & yes \\
S06 & -- & 376 & 40 & 26 & 83 & 4 & 0 & 0 & 0 &  \\
S07 & 498 & 117 & 119 & 12 & -- & 4 & 3 & 80 & 2 & yes \\
S08 & 10 & 1 & -- & 406 & 120 & 3 & 0 & 0 & 0 &  \\
S09 & -- & -- & 4 & -- & 78 & 2 & 0 & 0 & 0 &  \\
S10 & 20 & 6 & -- & -- & -- & 2 & 0 & 0 & 0 &  \\
S11 & 31 & 294 & -- & -- & 187 & 3 & 0 & 0 & 0 &  \\
S12 & 540 & 368 & 252 & 317 & 273 & 5 & 9 & 108 & 2 & yes \\
S13 & -- & 65 & 230 & -- & -- & 2 & 0 & 0 & 0 &  \\
S14 & 200 & 144 & 155 & 73 & 262 & 5 & 0 & 0 & 0 &  \\
S15 & 229 & -- & 150 & -- & -- & 2 & 0 & 0 & 0 &  \\
S16 & -- & 2 & 15 & 2 & 103 & 2 & 0 & 0 & 0 &  \\
S17 & 202 & 71 & 22 & 213 & 85 & 5 & 0 & 0 & 0 &  \\
S18 & 399 & 322 & 96 & 12 & 261 & 5 & 0 & 0 & 0 &  \\
S19 & 7 & 286 & 245 & -- & 38 & 4 & 5$^\dagger$ & 117 & 1 & yes \\
S20 & 121 & 213 & 95 & 237 & 238 & 5 & 0 & 0 & 0 &  \\
S21 & -- & 37 & -- & 3 & 44 & 3 & 3$^\dagger$ & 110 & 1 & yes \\
S22 & 2 & -- & -- & 50 & 2 & 1 & 7$^\dagger$ & 190 & 3 & yes \\
S23 & 100 & 129 & 44 & -- & 109 & 4 & 5$^\dagger$ & 94 & 3 & yes \\
S24 & 398 & 378 & 66 & 340 & 171 & 5 & 19 & 492 & 2 & yes \\
S25 & 87 & 348 & -- & 38 & 59 & 4 & 0 & 0 & 0 &  \\
S26 & -- & 324 & 400 & 683 & 105 & 4 & 0 & 0 & 0 &  \\
S27 & -- & 248 & 23 & 28 & 71 & 4 & 0 & 0 & 0 &  \\
S28 & 99 & 212 & -- & 146 & -- & 3 & 0 & 0 & 0 &  \\
S29 & 220 & 66 & -- & -- & 252 & 3 & 8 & 143 & 1 & yes \\
S30 & -- & 250 & -- & 480 & 54 & 3 & 4 & 57 & 1 & yes \\
S31 & 66 & 138 & 110 & 4 & 53 & 5 & 0 & 0 & 0 &  \\
S32 & 145 & 197 & 189 & 68 & 168 & 5 & 0 & 0 & 0 &  \\
S33 & 41 & 74 & 33 & 191 & 10 & 5 & 0 & 0 & 0 &  \\
S34 & -- & -- & -- & -- & -- & 0 & 0 & 0 & 1 &  \\
S35 & 156 & 18 & 124 & 561 & -- & 4 & 0 & 0 & 0 &  \\
S36 & 369 & 108 & 338 & 128 & 159 & 5 & 0 & 0 & 1 &  \\
S37 & 482 & -- & 95 & 203 & -- & 3 & 0 & 0 & 0 &  \\
S38 & 131 & -- & 143 & 246 & 178 & 4 & 0 & 0 & 0 &  \\
S39 & 155 & -- & -- & -- & 7 & 2 & 0 & 0 & 0 &  \\
S40 & -- & -- & -- & -- & -- & 0 & 0 & 0 & 0 &  \\
S41 & 290 & 264 & 84 & -- & 93 & 4 & 9 & 367 & 3 & yes \\
S42 & 281 & 188 & 283 & -- & 78 & 4 & 3 & 80 & 3 & yes \\
S43 & 23 & 50 & 2 & -- & 4 & 3 & 0 & 0 & 0 &  \\
S44 & 20 & 29 & 167 & 2 & 257 & 4 & 0 & 0 & 0 &  \\
S45 & 440 & 201 & -- & -- & 55 & 3 & 0 & 0 & 0 &  \\
S46 & 407 & 262 & 443 & 90 & -- & 4 & 0 & 0 & 0 &  \\
S47 & -- & -- & -- & 58 & 5 & 2 & 0 & 0 & 0 &  \\
S48 & 730 & 161 & -- & -- & 81 & 3 & 6$^\dagger$ & 128 & 3 & yes \\
S49 & -- & -- & -- & -- & 1 & 0 & 2$^\dagger$ & 71 & 1 & yes \\
S50 & 202 & 89 & 71 & -- & 55 & 4 & 3 & 69 & 1 & yes \\
S51 & -- & 75 & -- & 22 & 172 & 3 & 0 & 0 & 0 &  \\
S52 & 79 & 2 & -- & 182 & 168 & 3 & 7 & 150 & 1 & yes \\
S53 & 387 & 27 & 151 & 2 & -- & 3 & 0 & 0 & 0 &  \\
S54 & 385 & -- & 265 & 40 & 190 & 4 & 0 & 0 & 1 &  \\
S55 & 125 & 52 & 378 & 122 & 68 & 5 & 0 & 0 & 0 &  \\
S56 & 184 & -- & -- & -- & -- & 1 & 0 & 0 & 0 &  \\
S57 & 51 & 153 & 355 & 399 & -- & 4 & 0 & 0 & 0 &  \\
S58 & 433 & 134 & 77 & -- & 49 & 4 & 0 & 0 & 0 &  \\
\bottomrule\end{longtable}\end{footnotesize}}

\newcommand{\voicetable}{%
\begin{table}[t]\centering\small
\caption{The 16 learned voiceprints: anchors from instructor name calls, anchors from manual labels, and later comments matched to the print at $\geq$0.50 with a 0.05 margin.}
\label{tab:voices}
\begin{tabular}{lrrr}\toprule
Student & Called & Labelled & Matched\\\midrule
S05 & 1 & 0 & 8 \\
S07 & 2 & 0 & 1 \\
S12 & 1 & 0 & 8 \\
S19 & 1 & 0 & 4 \\
S21 & 1 & 0 & 2 \\
S22 & 1 & 0 & 6 \\
S23 & 1 & 0 & 4 \\
S24 & 2 & 0 & 17 \\
S29 & 1 & 0 & 7 \\
S30 & 1 & 0 & 3 \\
S41 & 2 & 0 & 7 \\
S42 & 2 & 0 & 1 \\
S48 & 2 & 0 & 4 \\
S49 & 1 & 0 & 1 \\
S50 & 1 & 0 & 2 \\
S52 & 1 & 0 & 6 \\
\bottomrule\end{tabular}\end{table}}

\title{\vspace{-1.2em}\Large\textbf{Lecture Capture as a Classroom Sensor:\\Automated Attendance and Participation from Existing Recordings\\in an MBA Course}}
\author{Abhinav Gupta\\\small UC Berkeley\\\small\href{mailto:abhinav.gupta@berkeley.edu}{\texttt{abhinav.gupta@berkeley.edu}} \and Tess Rubin\\\small UC Berkeley\\\small\href{mailto:tess_rubin@berkeley.edu}{\texttt{tess\_rubin@berkeley.edu}}}
\date{\small Code and data: \href{https://github.com/abhinavgupta621/classroom-sensor}{\texttt{github.com/abhinavgupta621/classroom-sensor}}\\\small Companion portal: \href{https://classroom-sensor.vercel.app}{\texttt{classroom-sensor.vercel.app}}\\\small Public edition: students appear as anonymous ids (S01--S58) and their faces are blurred.}

\begin{document}
\maketitle

\begin{abstract}
Business schools record nearly every class, yet attendance is still taken on paper and class participation is graded from memory. We ask how much of both can be recovered automatically from recordings that already exist. Using five two-hour sessions of one section (58 students) of a required MBA leadership course, captured by the school's Panopto lecture-capture system, we built an open pipeline that (i) finds when the moving room camera shows students, (ii) recognises faces against seating-chart photographs, (iii) transcribes speech with the class list as hints, (iv) separates the instructor's voice, and (v) names student voices by pairing the instructor's name calls with the voice that answers, then carrying those voiceprints across classes. Matching every face against all four sections of the course lets us count identification errors without hand labels. At 1080p, 0 of 1{,}393 named faces were certainly wrong; at the more common 576p, a 3-second presence rule held certain errors to 3 across five classes ($\approx$2\% estimated). Across the five classes 55 of 58 students were recognised at least once, but the camera showed the class only 11--33\% of the time, so the system establishes who was present, not who was absent. Reading three frames per student shot found 79--86\% of the students a full scan finds, in 44--61\,s. Voiceprints seeded by name calls separated speakers well (median similarity 0.70 for the same student, 0.09 for different students) and named the right student in every held-out test (4 of 4 with a face check, 6 of 6 without); no visual cue (gaze, mouth motion, centrality, size) could pick the speaker among a median of 16 faces. The core steps take about 10 minutes per class on one laptop, most of it transcription. We also document data-quality problems in production lecture capture and the infrastructure that would close the remaining gaps.
\end{abstract}

\section{Introduction}
The Haas student body was asked to automate three classroom chores: attendance (a paper sheet passed around the room, slow and easy to fill in for a friend), contribution (who spoke, how often and how well) and engagement (who is drifting out of the discussion). The constraint was to use infrastructure the school already runs: Panopto lecture capture in every classroom, a pan--tilt--zoom (PTZ) camera that turns toward a student when they press a desk microphone, ceiling and lectern microphones, course rosters and seating-chart headshots. The administration collected student consent this year for face recognition, voiceprints and all-party recording, so we could study real recordings. Our question was what existing lecture capture can tell an instructor, how accurately and how quickly, and what it misses.

\begin{figure}[t]\centering
\includegraphics[width=\figw]{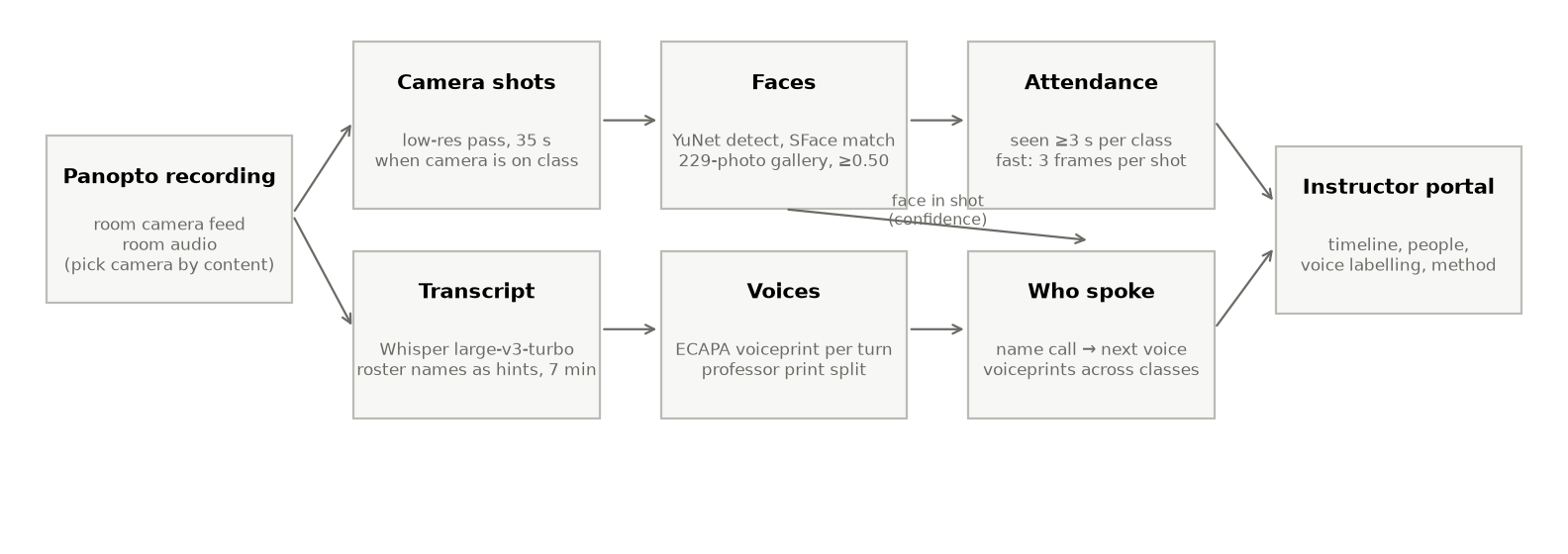}
\caption{Pipeline. Everything runs after class; results appear in a local instructor portal.}\label{fig:pipeline}
\end{figure}

\paragraph{Contributions.} (1) An end-to-end, open pipeline that runs after class on a single laptop (Fig.~\ref{fig:pipeline}). (2) Error measurement without hand labels, by matching against all four sections of the course. (3) Positive and negative results on speaker identification, including a voiceprint bootstrapping method seeded by instructor name calls. (4) A catalogue of data-quality problems in production lecture capture. (5) A sizing of what it would take to make the system complete.

\section{Related work}
Automated attendance by face recognition has been demonstrated in controlled settings with frontal, close-range cameras; classroom deployments report sharp losses with distance, pose and occlusion. We use lightweight, permissively licensed models, YuNet for detection \cite{yunet} and SFace for recognition \cite{sface}, and measure them on a production PTZ camera rather than a purpose-mounted one. For speech we use Whisper \cite{whisper} and ECAPA-TDNN speaker embeddings \cite{ecapa,speechbrain}. Audio-visual active-speaker detection (e.g.\ TalkNet \cite{talknet}) reaches high accuracy on close faces but degrades for faces under 64\,px, which is where our classroom sits. Discourse-level measures of teaching, such as conversational uptake \cite{uptake}, motivate the transcript analyses. Emotion recognition in education is prohibited in the EU \cite{aiact}; we neither attempt nor recommend it.

\section{Setting and data}
\paragraph{Course and room.} One section (58 students) of a first-year MBA core course on leading people, taught by one instructor in a tiered classroom, observed in five sessions between 22 September and 8 October 2026 (each $\approx$120\,min of recording). Sessions mix lecture, cold-called case discussion, a quiz game and breaks. Table~\sessiontable\ref{tab:sessions} summarises them.

\paragraph{Recordings.} Each Panopto session holds a slides feed and a room-camera feed with room audio. The camera moves between presets (front of room and several views of the student tiers, Fig.~\ref{fig:views}). One session was stored at 1920$\times$1080; four were stored at 1024$\times$576. The stream labelled as the camera was in fact the slides in four of five sessions (Fig.~\ref{fig:swap}).

\paragraph{Reference faces.} 230 seating-chart headshots across all four sections of the course (229 usable). Matching against every section is what makes errors countable: a face named as a student of another section is certainly wrong.

\paragraph{Ground truth.} Section membership; desk name cards; names the instructor said aloud; a model reading the transcript in context; two repeated scoring runs of the same model. There was no attendance register, which is the main gap.

\begin{figure*}[t]\centering
\includegraphics[width=\textwidth]{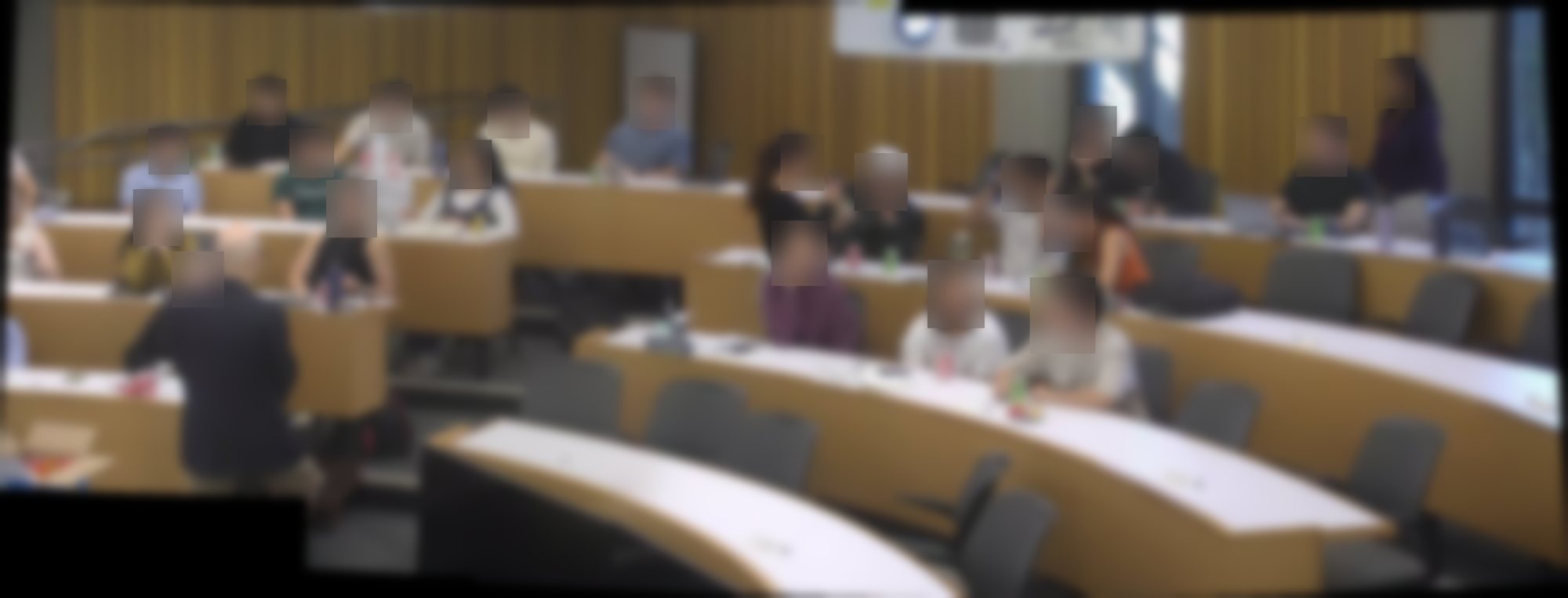}
\caption{Part of the classroom, stitched from three overlapping camera presets on 8 October. Only presets that overlap cleanly can be joined; the rest are shown in Fig.~\ref{fig:views}.}\label{fig:pano}
\end{figure*}

\section{Method}
\subsection{Video}
\begin{figure*}[t]\centering
\includegraphics[width=0.9\textwidth]{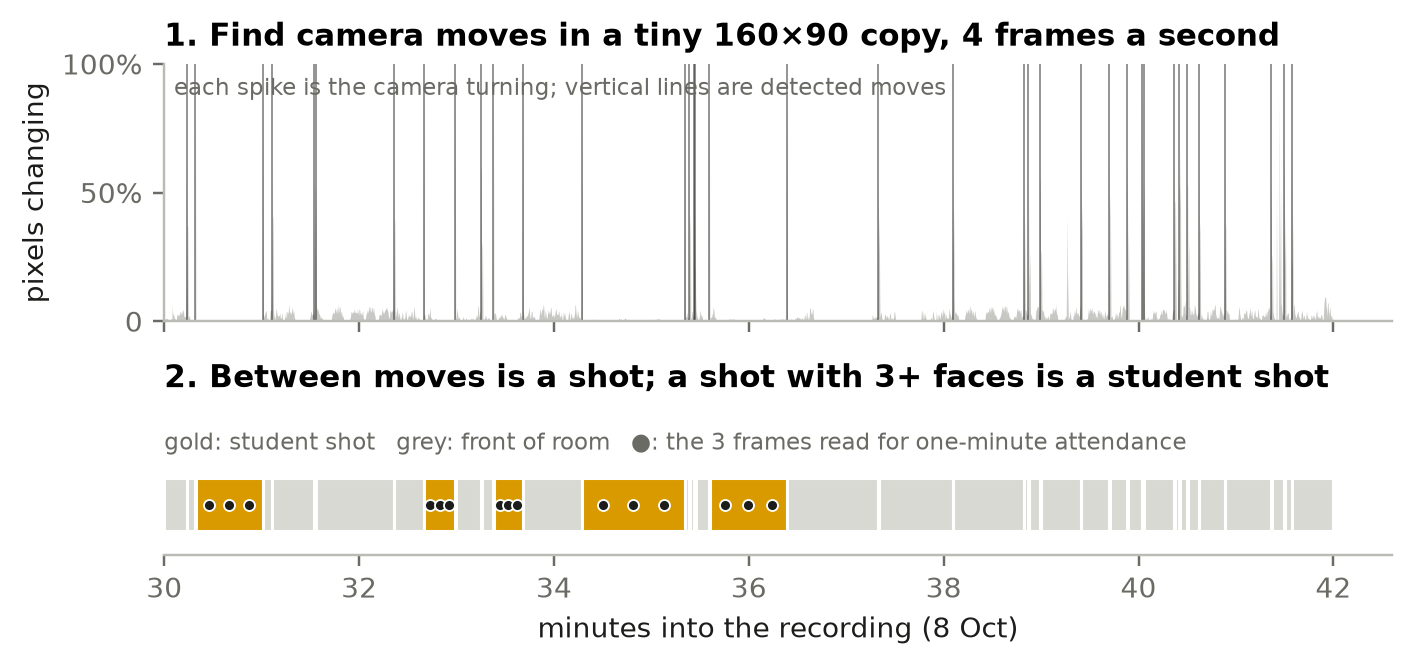}
\caption{How camera shots are found (twelve minutes of 8 October). The share of pixels changing between frames spikes when the camera turns; the stills between turns are shots. Student shots (gold) are where faces are read; the one-minute pass reads only the three marked frames in each.}\label{fig:howshots}
\end{figure*}

\begin{figure}[t]\centering
\includegraphics[width=0.95\figw]{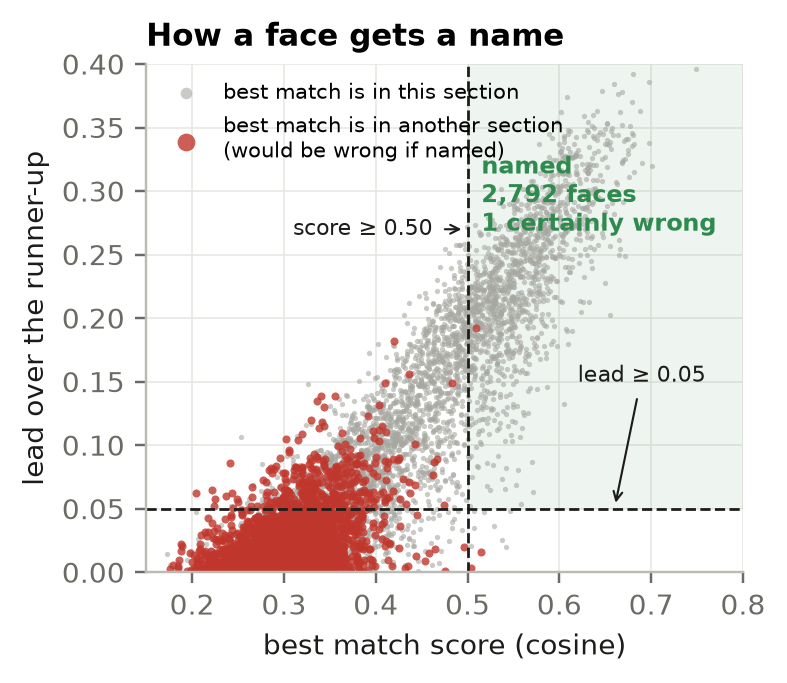}
\caption{How a face gets a name, for every face detected at 1080p on 8 October (grey points subsampled). A face is named only in the shaded corner: score $\geq$0.50 and a lead of $\geq$0.05 over the next-best student. Red faces' best match is a student from another section, so naming them would certainly be wrong; the rule rejects all but one. Table~\ref{tab:id} uses a 2-second sample of the same class.}\label{fig:howfaces}
\end{figure}

A low-resolution pass (160$\times$90 grey, 4\,fps) over the whole recording finds camera moves; stills between moves are shots, grouped into presets by appearance, and a shot with $\geq$3 faces is a student shot. Faces $\geq$24\,px are detected with YuNet and embedded with SFace (128-d). A face is named when its best cosine similarity to the gallery is $\geq\tau{=}0.50$ and exceeds the runner-up by $\geq$0.05 (Fig.~\ref{fig:howfaces}). Faces are followed through a shot by box overlap; a student is \emph{present} in a class when named for $\geq$3\,s. Figure~\ref{fig:howshots} shows how shots are found and which frames the one-minute attendance pass reads.

\subsection{Audio}
Audio is transcribed with Whisper large-v3-turbo (mlx on Apple Silicon) with the section's first names and course terms as prompt. Words form turns at pauses of 0.8\,s; each turn of $\geq$1.5\,s gets an ECAPA voiceprint. The instructor print is the largest voice cluster of one session; it transfers across sessions (cosine 0.96--0.97).

\subsection{Naming voices}
\begin{figure*}[t]\centering
\includegraphics[width=0.9\textwidth]{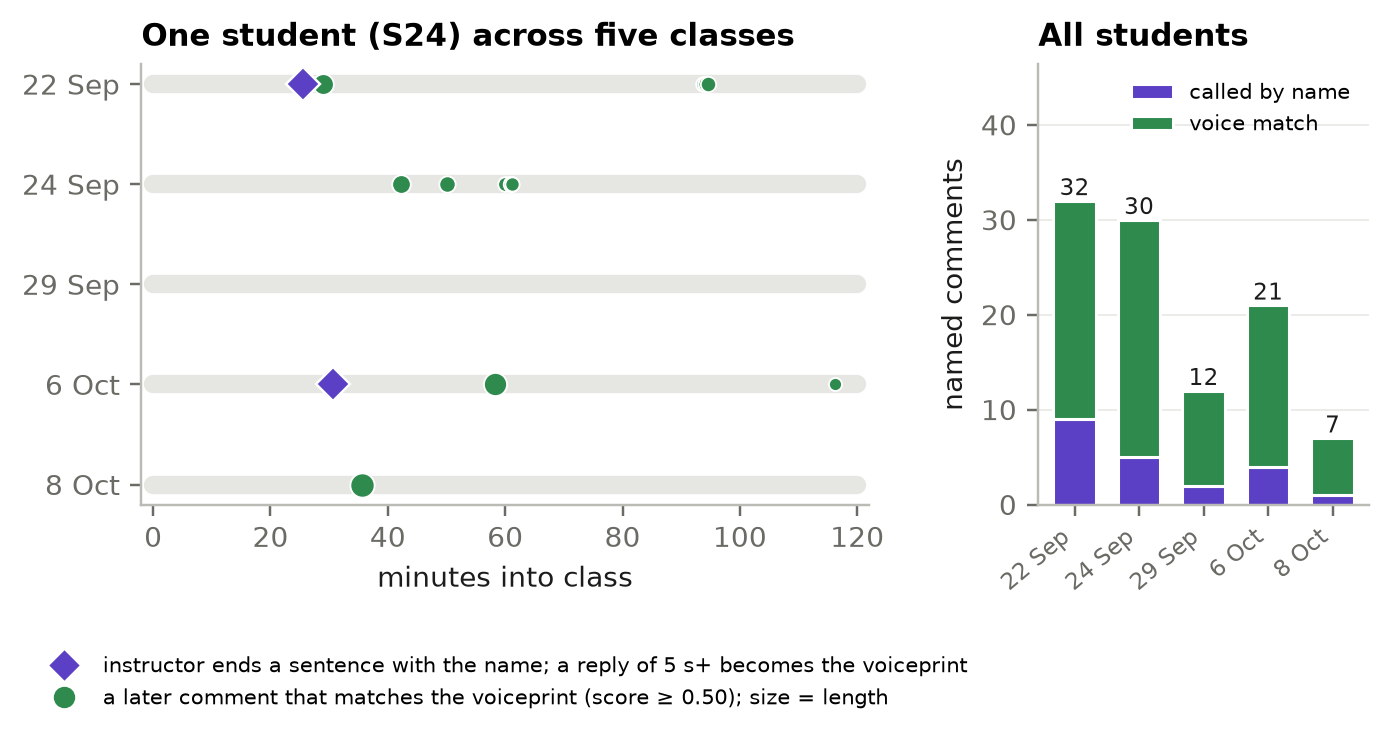}
\caption{How a voice gets a name. Left: one student, anonymised. Two name calls (diamonds) gave the voiceprint; every other comment (circles) was matched to it by voice, including in classes where the student was never called on. Right: named comments per class for all students, by method.}\label{fig:howvoices}
\end{figure*}

When the instructor ends a sentence with a student's first name (``\ldots[name], go.'') and a non-instructor turn of $\geq$5\,s starts within 20\,s, that turn is labelled with the student. Labelled turns form the student's voiceprint; other student turns in any class are named when they match a print at $\geq$0.50 with a 0.05 margin. For evaluation we also required the called student's face on camera during the reply; the portal drops that check to learn more voices. A portal lets a person label unknown voice clusters, which become anchors. Figure~\ref{fig:howvoices} follows one student through the five classes.

\subsection{Evaluation without hand labels}
Identification errors are certain cross-section matches, scaled by 229/172 to estimate within-section errors. Verification uses same-seat pairs one second apart (same person) and same-frame pairs (different people). Voice accuracy is leave-one-class-out: each student's print is built from their anchors in the other classes and tested on their anchors in the held-out class.

\section{Results}
\subsection{Naming a visible face is precise}
\begin{figure}[t]\centering
\includegraphics[width=\figw]{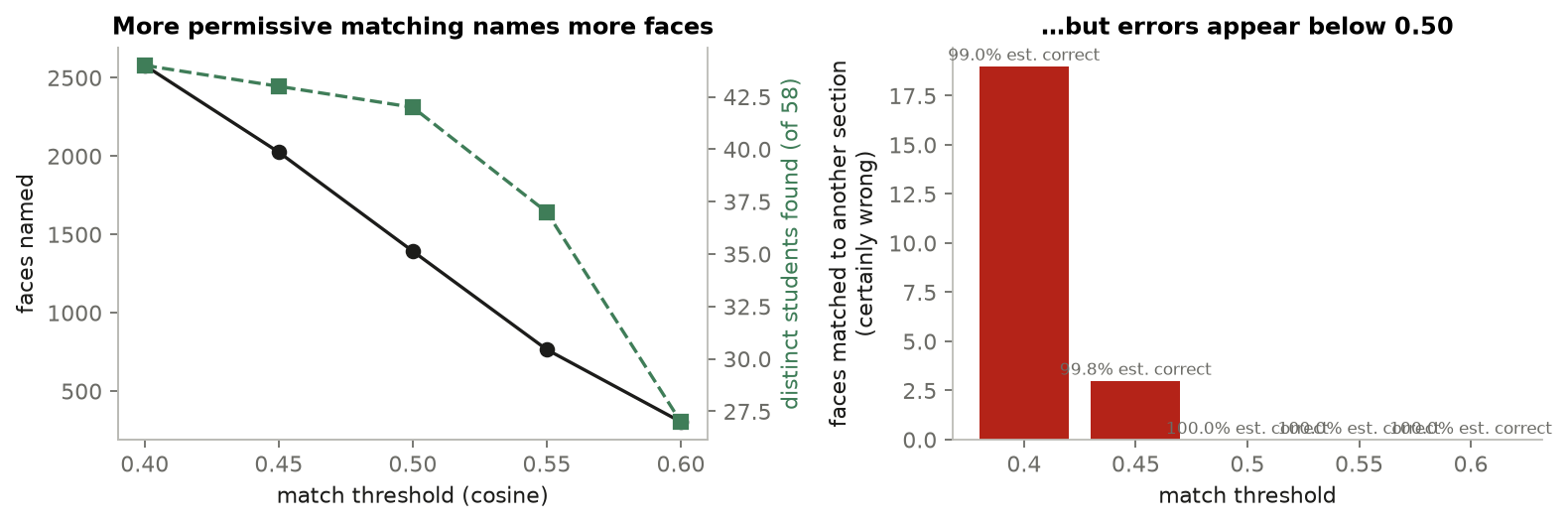}
\caption{Faces named, students found and certain errors by threshold (1080p).}\label{fig:id}
\end{figure}

\begin{figure}[t]\centering
\includegraphics[width=0.92\figw]{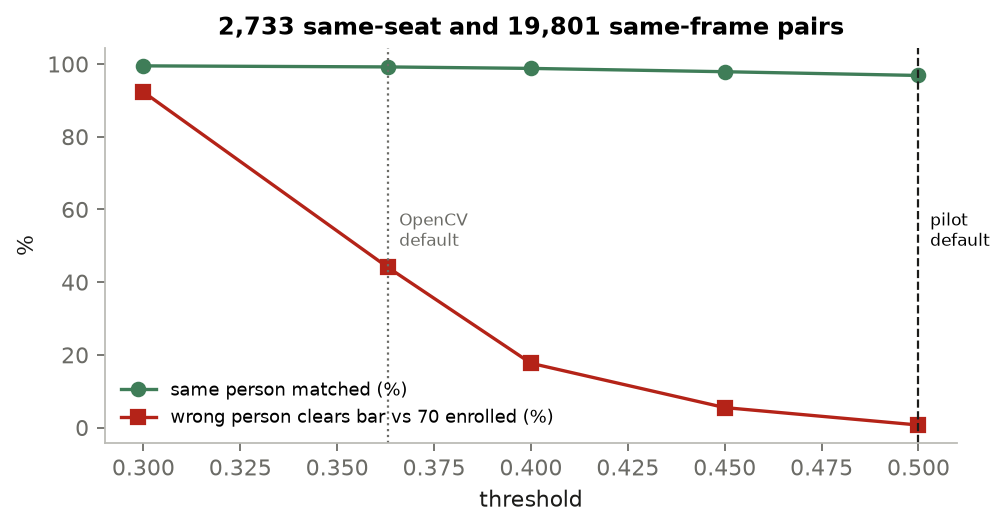}
\caption{Same-person match rate and false-match rate against a 70-person gallery.}\label{fig:ver}
\end{figure}

Table~\idtable\ref{tab:id} and Fig.~\ref{fig:id} show the identification trade-off on the 1080p session. At $\tau{=}0.50$, 1{,}393 faces were named with no certain error (estimated error below $\approx$0.3\%, 95\% confidence); at 0.40 there were 19. The same seat a second later received the same name in 744 of 744 cases. Verification on the room's own footage (Table~\vertable\ref{tab:ver}, Fig.~\ref{fig:ver}) shows why the library default of 0.363 is unsafe for a full class: an unknown face would clear the bar against at least one of 70 students 44\% of the time, against 0.7\% at 0.50.


\begin{figure*}[p]\centering
\includegraphics[width=\textwidth]{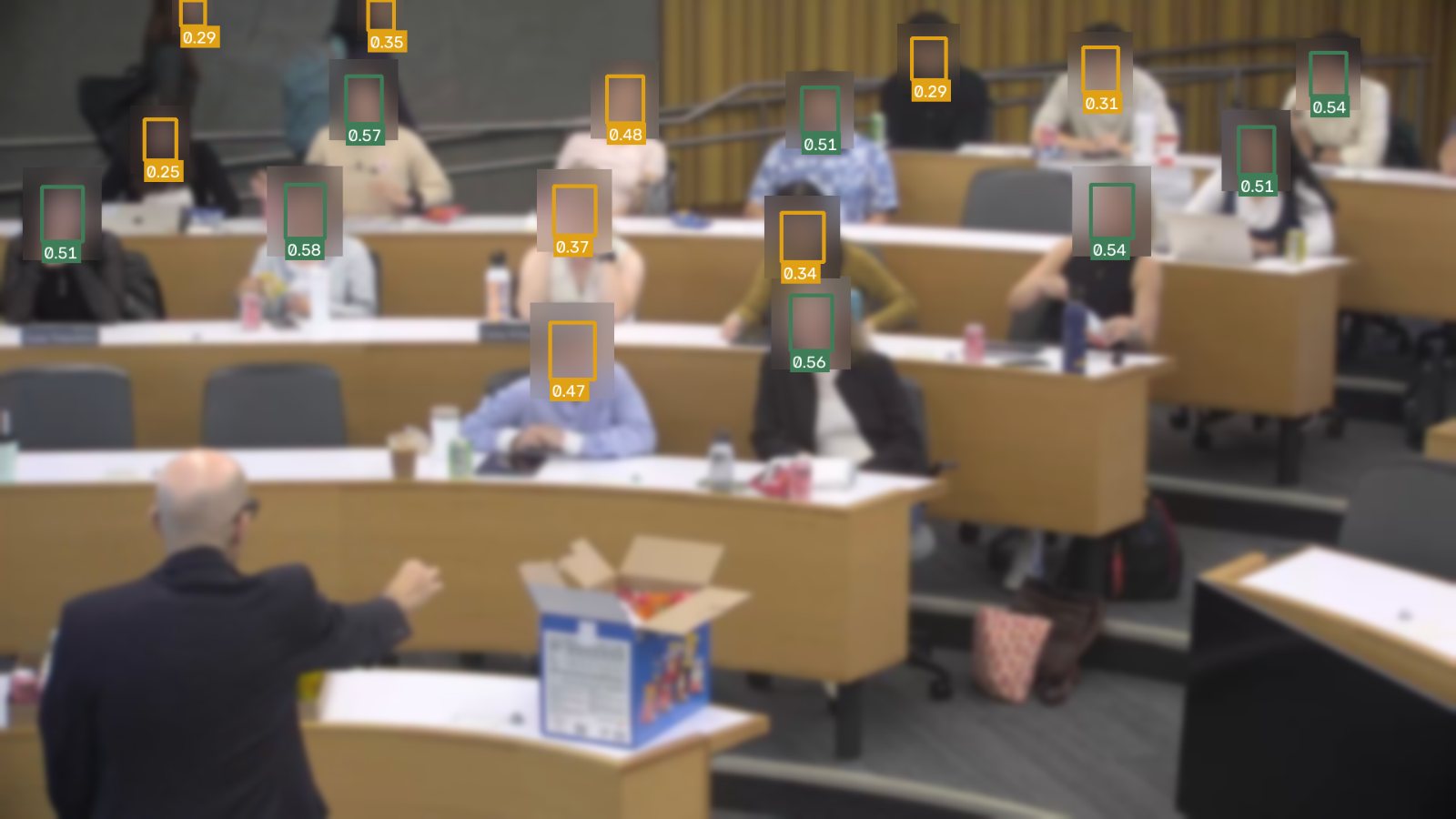}
\caption{Busiest student shot, 8 October (1920$\times$1080). Green: named at $\geq$0.50 with margin. Amber: best score, left unnamed. Faces blurred.}\label{fig:tag8}
\vspace{0.8em}
\includegraphics[width=\textwidth]{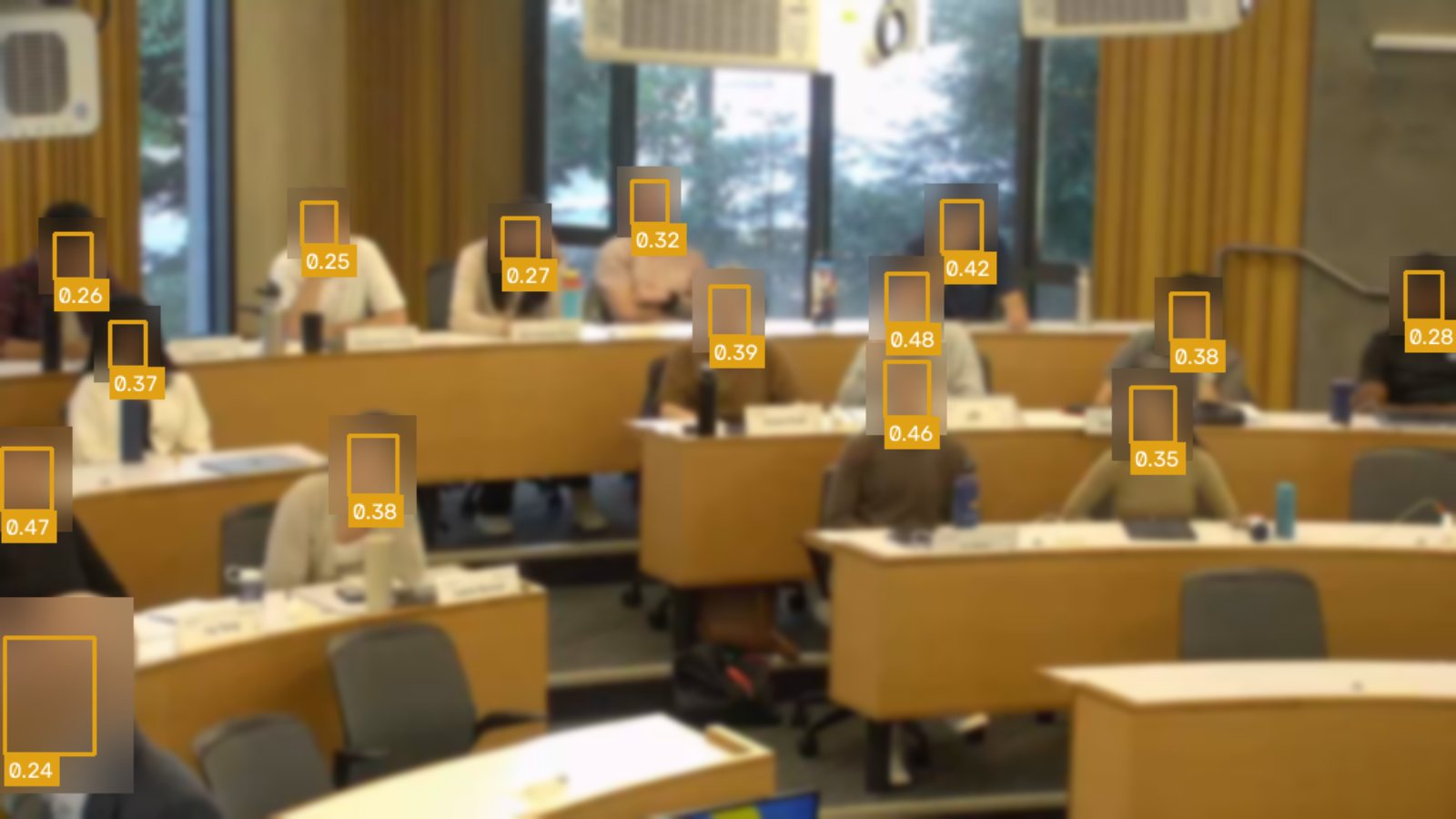}
\caption{Busiest student shot, 22 September (1024$\times$576). Faces are about half the size; scores fall and more faces are left unnamed. Faces blurred.}\label{fig:tag22}
\end{figure*}

Figures~\ref{fig:tag8} and \ref{fig:tag22} show every detected face in the busiest student shot of two classes with its best match score (faces blurred in this edition). In Fig.~\ref{fig:tag8}, one face named at 0.51 sits directly above a desk card with the same name, an independent confirmation. Amber faces scored below the threshold or margin and stay unnamed.

\paragraph{Name cards.} On-device text recognition on zoomed tiles read 309 desk cards in 298 student shots; every reading was a real section name. Only 5 of 58 students had a card the camera could read, and the card agreed with the face above it in 15 of 22 links; all 7 disagreements were one card linked to the adjacent student.

\subsection{Camera coverage limits attendance}
\begin{figure*}[t]\centering
\includegraphics[width=\textwidth]{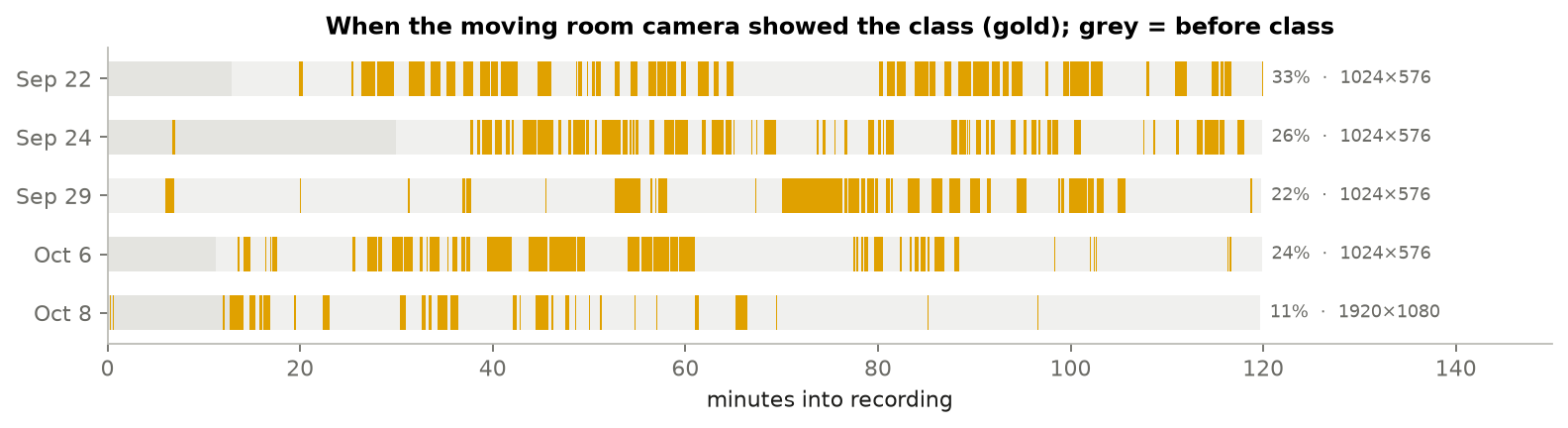}
\caption{When the room camera showed the class (gold). Grey: before class (music, empty room).}\label{fig:cov}
\end{figure*}

\begin{figure}[t]\centering
\includegraphics[width=\figw]{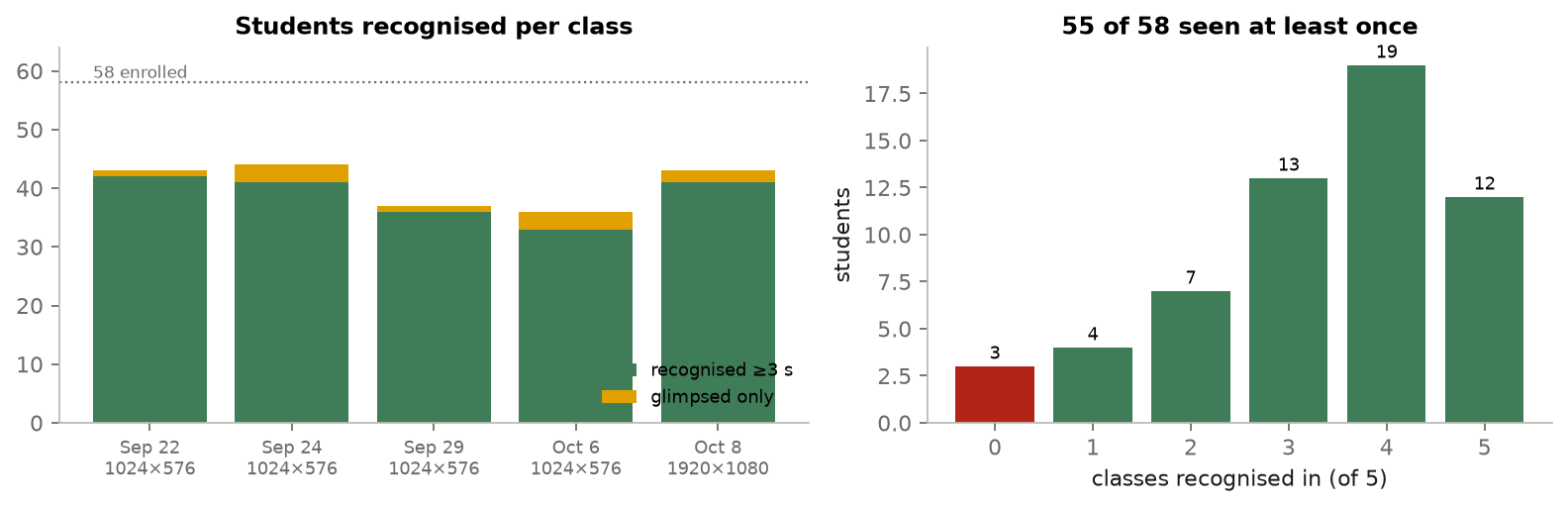}
\caption{Left: recognised per class. Right: students by number of classes recognised.}\label{fig:att}
\end{figure}

The camera serves remote viewers: it follows the instructor and the active microphone. Student shots covered 22--33\% of discussion classes and 11\% of the quiz-game class (Fig.~\ref{fig:cov}). Per class the system named 33--42 of 58 students for $\geq$3\,s (Fig.~\ref{fig:att}); with that rule, certain errors across all five classes fell from 10 to 3. Counting faces seen but not named by clustering was unreliable (on 8 October it estimated 86 unnamed people in a 58-student section), so absence cannot be inferred. Across classes coverage accumulates: 55 of 58 were recognised at least once and 31 in four or five classes (Table~\ref{tab:students}, Fig.~\ref{fig:attname}).


\subsection{Attendance in a minute}
\begin{figure}[t]\centering
\includegraphics[width=0.92\figw]{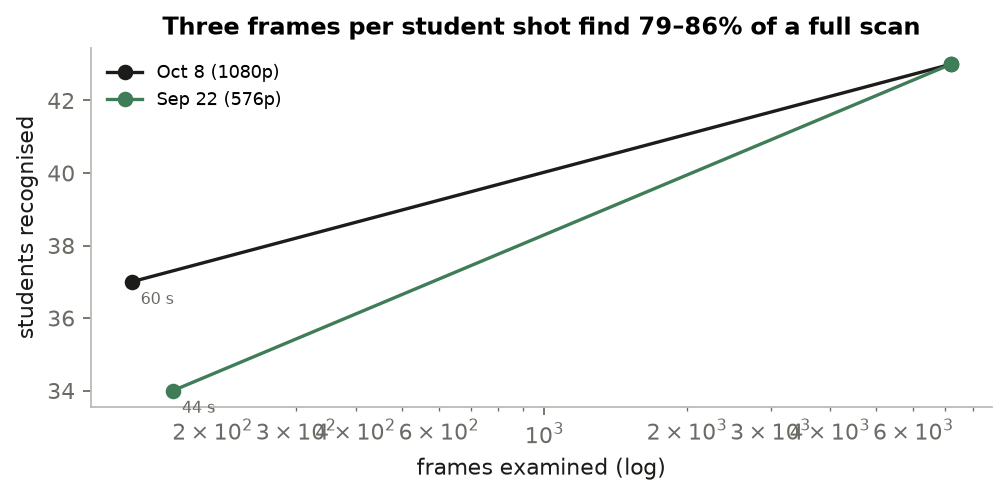}
\caption{Students recognised versus frames examined.}\label{fig:fast}
\end{figure}

Reading three frames from each student shot (135--165 frames rather than 7{,}200) found 79--86\% of the students a full scan finds (34 of 43 on 22 September, 37 of 43 on 8 October) in 44--61\,s end to end, with one certain error at 576p and none at 1080p; most of the time is the shot-finding pass (Fig.~\ref{fig:fast}).

\subsection{Speech}
\begin{figure}[t]\centering
\includegraphics[width=\figw]{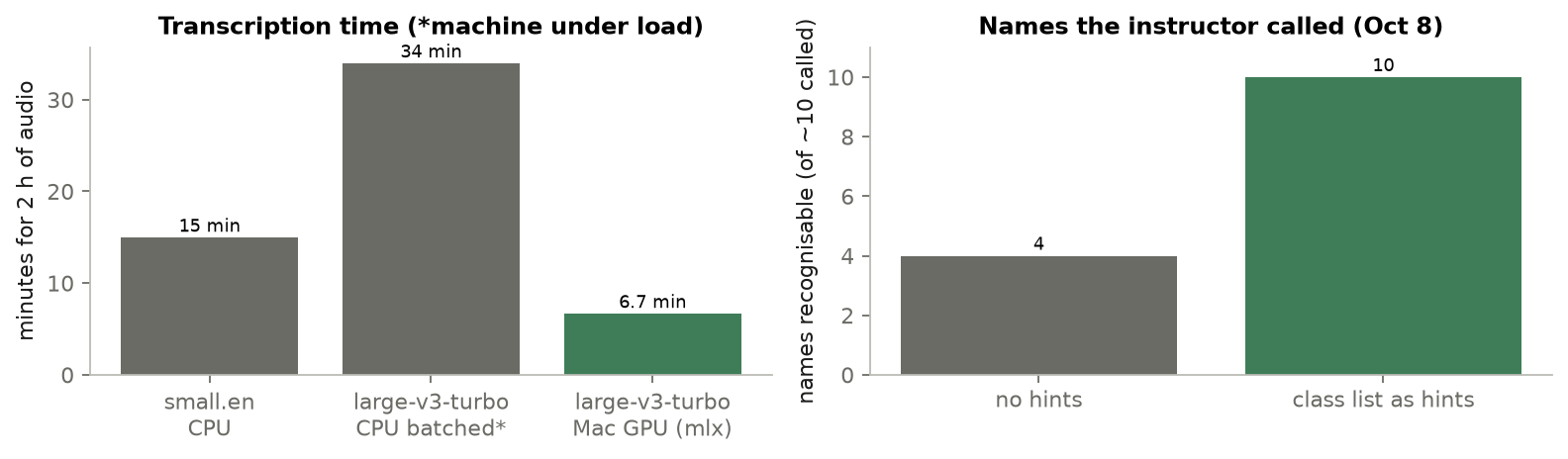}
\caption{Transcription time and recognisable called names.}\label{fig:asr}
\end{figure}

Instructor speech was $\approx$95\% clean; student points were recoverable in most turns. With the class list as a prompt, 9--22 section first names appeared in each class's transcript; in an exploratory comparison on one class, recognisable called names rose from 4 to 9--10 of 10. Large-v3-turbo on the laptop GPU transcribed two hours in 6.5--8.0\,min in four classes and 12.5\,min in the fifth (Fig.~\ref{fig:asr}). Pause-split turns blur speakers: a hand check of one class found turns labelled ``student'' that were the instructor, music or the quiz scorekeeper, and instructor turns that contained student speech; scanning 2\,s windows against the instructor print recovered most of the hidden student segments. These two checks were exploratory and are not in the released data.

\subsection{Identifying the speaker}
\begin{figure}[t]\centering
\includegraphics[width=0.92\figw]{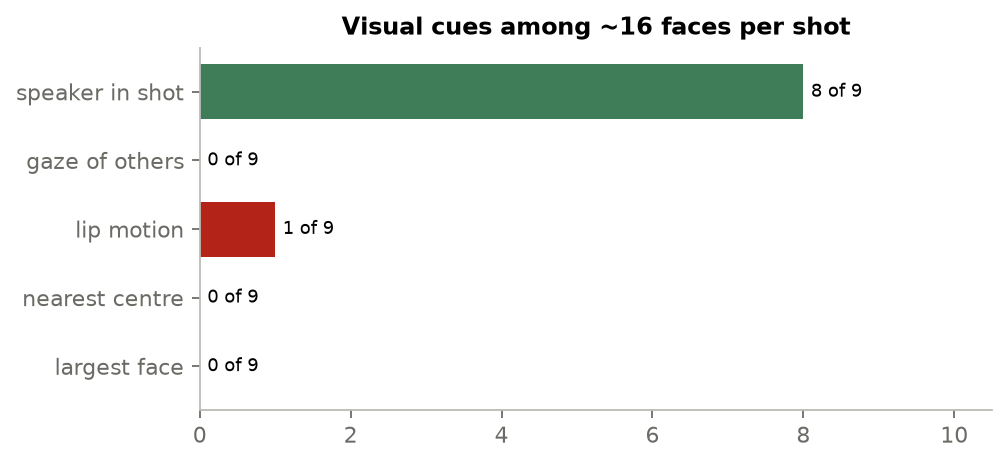}
\caption{Visual cues against known speakers.}\label{fig:cues}
\end{figure}

The camera follows the microphone: all 15 genuine student comments in one class had the camera on students during or just before them, and a student shot began within 16\,s of a comment three times more often than chance. The speaker was in shot for 8 of 9 known comments, but among a median of 16 faces (9 named), gaze of others picked the speaker 0 times, mouth motion once, nearest-centre 0 and largest face 0 (Fig.~\ref{fig:cues}).

\begin{figure}[t]\centering
\includegraphics[width=0.92\figw]{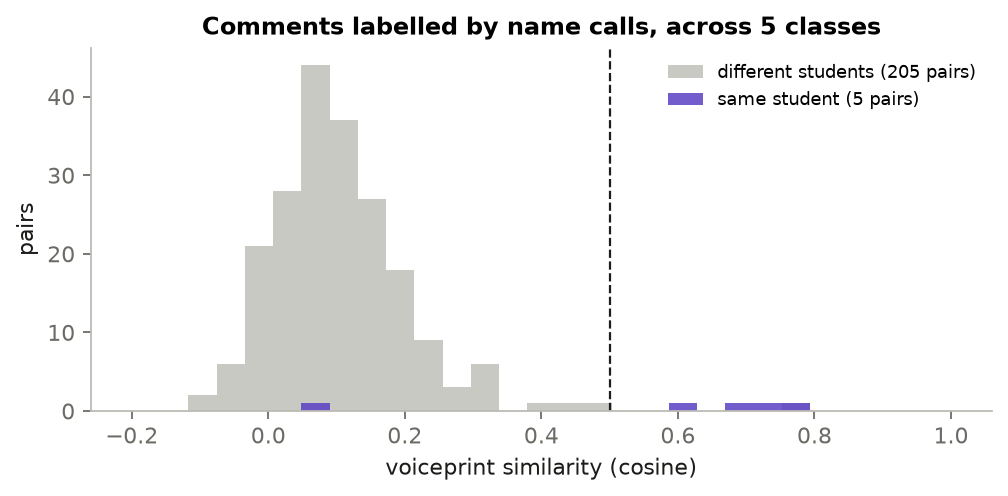}
\caption{Pairwise voiceprint similarity of name-call-labelled comments.}\label{fig:voice}
\end{figure}

Voices separate well: across 210 pairs of name-call-labelled comments, the median similarity was 0.70 for the same student and 0.09 for different students (Fig.~\ref{fig:voice}). The single same-student pair at 0.07 exposed a mislabelled anchor, which is how errors in this method surface. With the face check, the evaluation found 14 anchors for 10 students; in leave-one-class-out tests each print named its student correctly in the held-out class 4 of 4 times (similarity 0.70--0.78, lead over the next student 0.47--0.59). An earlier, looser rule (any mention of a name, not only at the end of a sentence) labelled replies wrongly, which is why the rule asks for the name to end the instructor's sentence. Without the face check, the portal's rule finds 21 anchors and learns 16 voiceprints (Table~\voicetable\ref{tab:voices}); the same test is correct 6 of 6 times, and its prints name 14--30\% of student-side audio per class, against 6--22\% with the face check. In the portal 102 comments are named (81 by voice match); 16 of those voice matches come from classes in which the camera never recognised that student, and the portal flags them for review. Every anchor, held-out test and named comment is in the released data (\texttt{voice\_anchors.csv}, \texttt{voice\_heldout\_tests.csv}, \texttt{voice\_matches.csv}).

\begin{figure*}[t]\centering
\includegraphics[width=\textwidth]{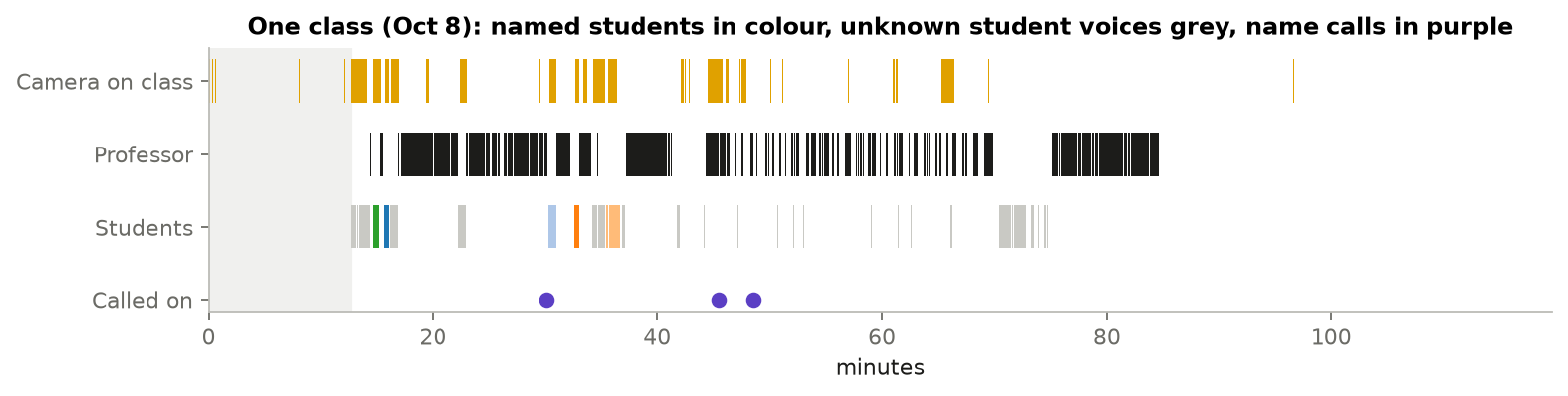}
\caption{One class on a timeline (8 October): camera on students, instructor speech, student speech (named in colour, unknown in grey) and name calls.}\label{fig:timeline}
\end{figure*}

\subsection{Scoring contributions}
\begin{figure}[t]\centering
\includegraphics[width=0.72\figw]{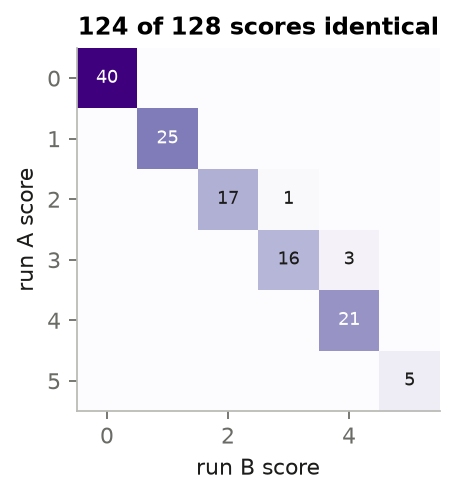}
\caption{Agreement between two repeated scoring runs of the same model and prompt.}\label{fig:rubric}
\end{figure}

Two repeated runs of the same Claude model and prompt scored each student turn of one class on four 1--5 criteria (relevance, evidence, building on others, insight). Scores were identical in 124 of 128 cases and never differed by more than one point (Fig.~\ref{fig:rubric}); both runs flagged the same 12 turns as non-student audio. No instructor has yet scored a blind sample, so validity is untested. From the transcript alone the model also produced a class-phase timeline, a discussion summary, the questions students asked, an exam review sheet with all 20 quiz questions and follow-up email points.

\subsection{Speed and cost}
\begin{figure}[t]\centering
\includegraphics[width=0.92\figw]{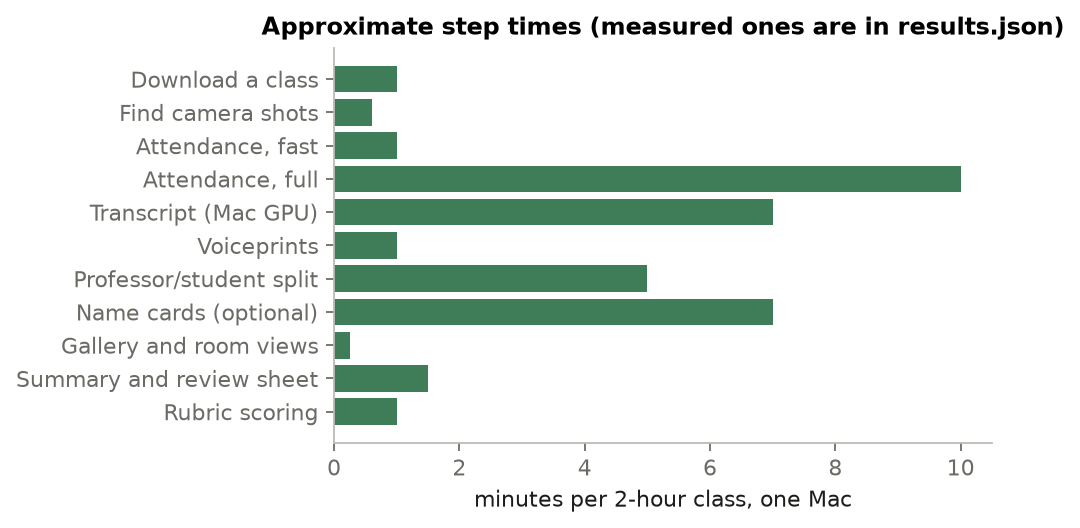}
\caption{Step times per class on one Mac.}\label{fig:time}
\end{figure}

Per two-hour class on one laptop, measured: shot finding 27--44\,s, fast attendance 44--61\,s, transcript 6.5--12.5\,min; estimated: download and voiceprints about 1\,min each (Fig.~\ref{fig:time}; measured times are in \texttt{timings.csv}). On a cloud GPU we estimate $\approx$\$0.10 per class.

\section{Discussion}
\paragraph{What worked.} Naming a visible face; term-level presence; one-minute attendance; transcripts with names; naming a voice after a name call and carrying it across classes; transcript summaries and review sheets.
\paragraph{What partly worked.} Absence (coverage-limited); naming most student talk (6--30\% of it after five classes, depending on the rule); instructor/student separation at turn boundaries; contribution scores (consistent, unvalidated).
\paragraph{What failed.} Picking the speaker from video at 30--60\,px; late arrival and early exit; name cards as an attendance source.

\begin{figure}[t]\centering
\includegraphics[width=\figw]{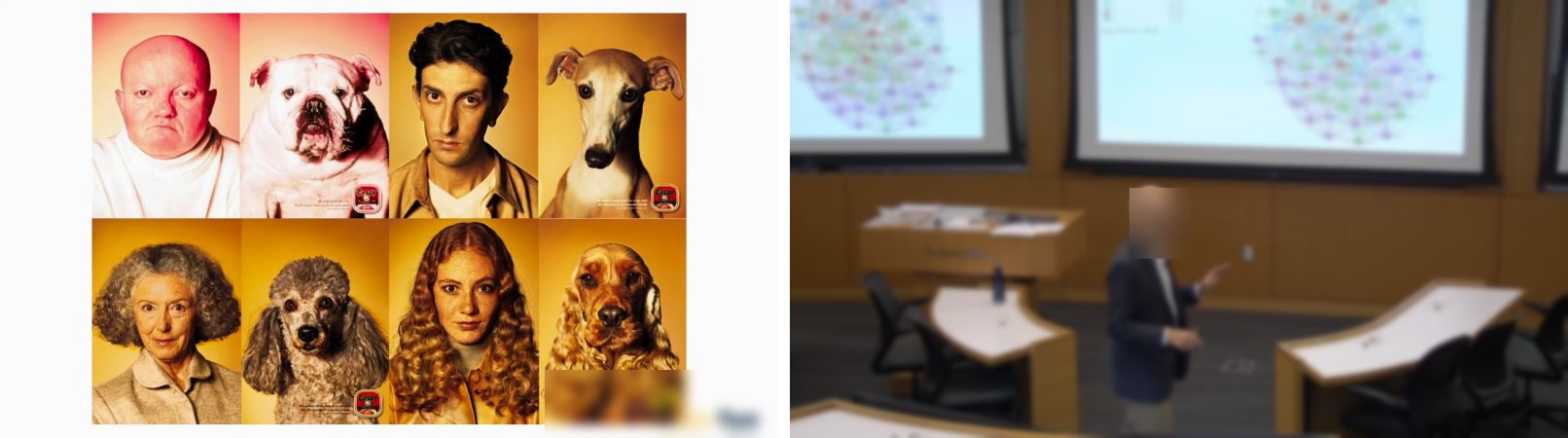}
\caption{29 September: the stream Panopto labelled as the room camera (left) is the slides; the other stream (right) is the camera.}\label{fig:swap}
\end{figure}

\paragraph{Data-quality traps.} (1) Feed labels swap between sessions (Fig.~\ref{fig:swap}); identify the camera by content. (2) Resolution varies (1080p vs 576p) and halves face size; keep originals. (3) Recordings start early (once by 30\,min) and include music, which transcribes as student speech. (4) Guests and the quiz scorekeeper look like students until labelled. (5) The PTZ camera is aimed for remote viewers, so it films quiet students least.

\section{Limitations}
One section, one room, one instructor, five sessions. No attendance register, so attendance recall is unmeasured. Voice results rest on 10--16 speakers and four to six held-out tests. Error estimates assume errors spread evenly across the gallery. Contribution scores are not validated against the instructor, and we did not test for differences in error by accent, gender or skin tone, which must precede any use affecting students.

\section{What it would take}
A teaching assistant's tick sheet for three to five classes (true recall); the room controller's log of which desk microphone was pressed (speaker seat, likely free); keeping original-resolution camera files; a fixed wide camera on the student tiers (\$1.4--4.3k per room); a ceiling microphone array with talker position if no microphone log exists (\$4--7.6k); Panopto administrative API access; an instructor's blind scores on 20--30 comments; and P4-approved storage with a GPU host ($\approx$\$0.10 per class).

\section{Ethics and data handling}
Students consented to face recognition, voiceprints and recording, and the administration approved this research use. Face templates are P4 data under UC Berkeley's classification and per-student records are FERPA education records. Recordings, templates and voiceprints remain on the pilot machine; the identified results stay there too. This public edition replaces student names with anonymous ids and blurs student faces; the companion portal uses pseudonyms and has no audio or transcript text. We did not infer emotion or attention, rank students publicly or grade automatically.

\section{Conclusion}
Lecture capture that schools already run can tell an instructor who was in the room and what was said. Face naming was precise, transcripts were accurate once the model had the class list, and the instructor's habit of calling students by name was enough to learn their voices. The remaining gaps come from where the camera points and from not knowing which desk microphone was pressed. A microphone log, a fixed wide camera on the student tiers and a week of hand-taken attendance would close most of them.

\onecolumn
\appendix
\section{Every camera view of the class (faces blurred)}
\begin{figure}[H]\centering
\includegraphics[width=0.85\textwidth]{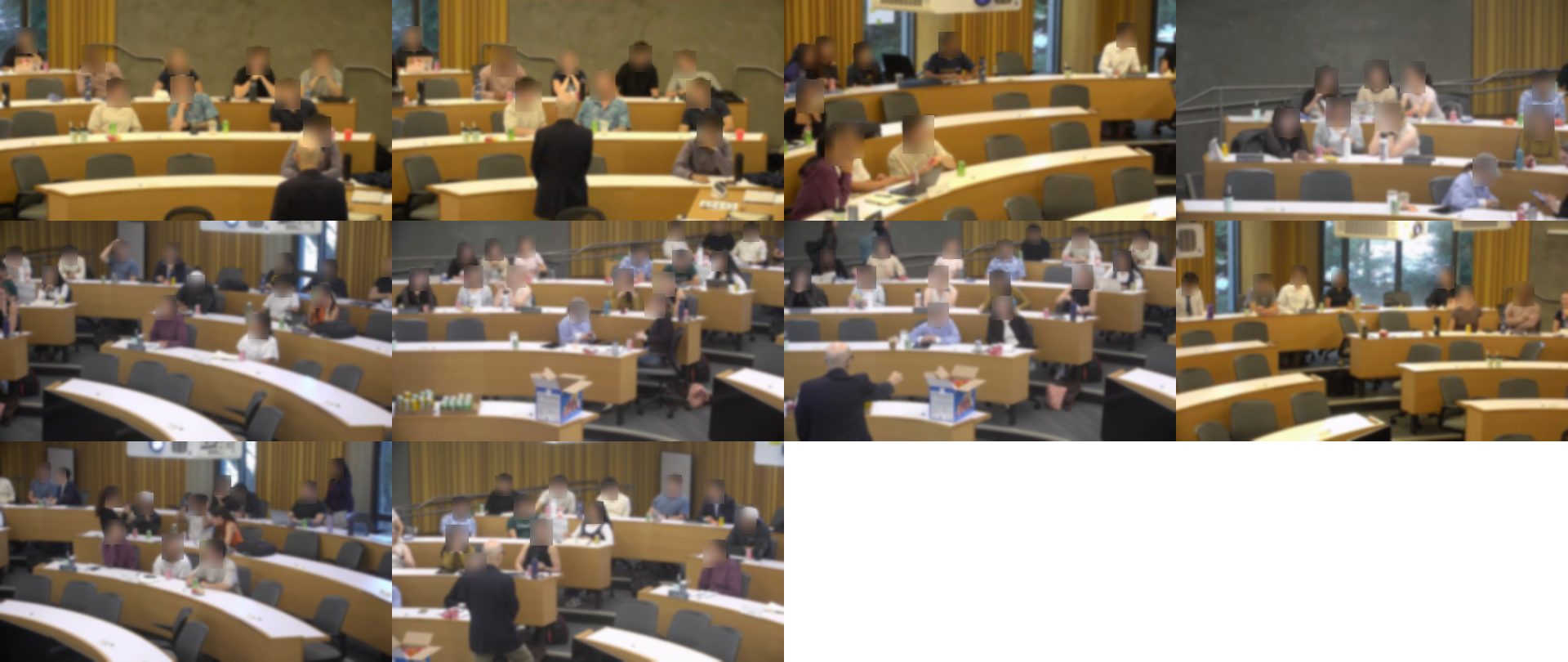}
\caption{The clearest frame from each camera preset that showed students on 8 October.}\label{fig:views}
\end{figure}

\section{Student shots from the other classes (faces blurred)}
\begin{figure}[H]\centering
\includegraphics[width=0.49\textwidth]{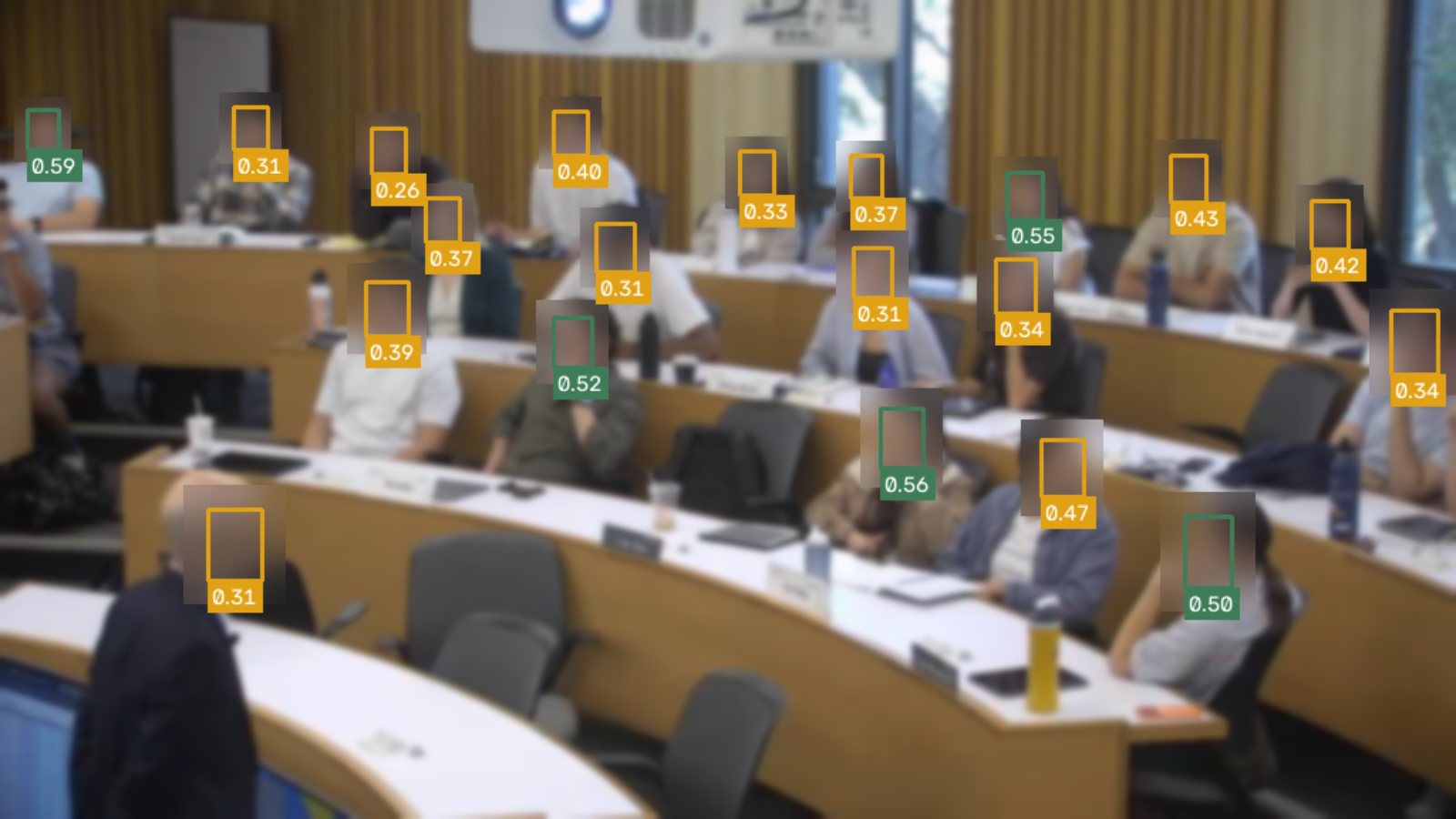}\hfill
\includegraphics[width=0.49\textwidth]{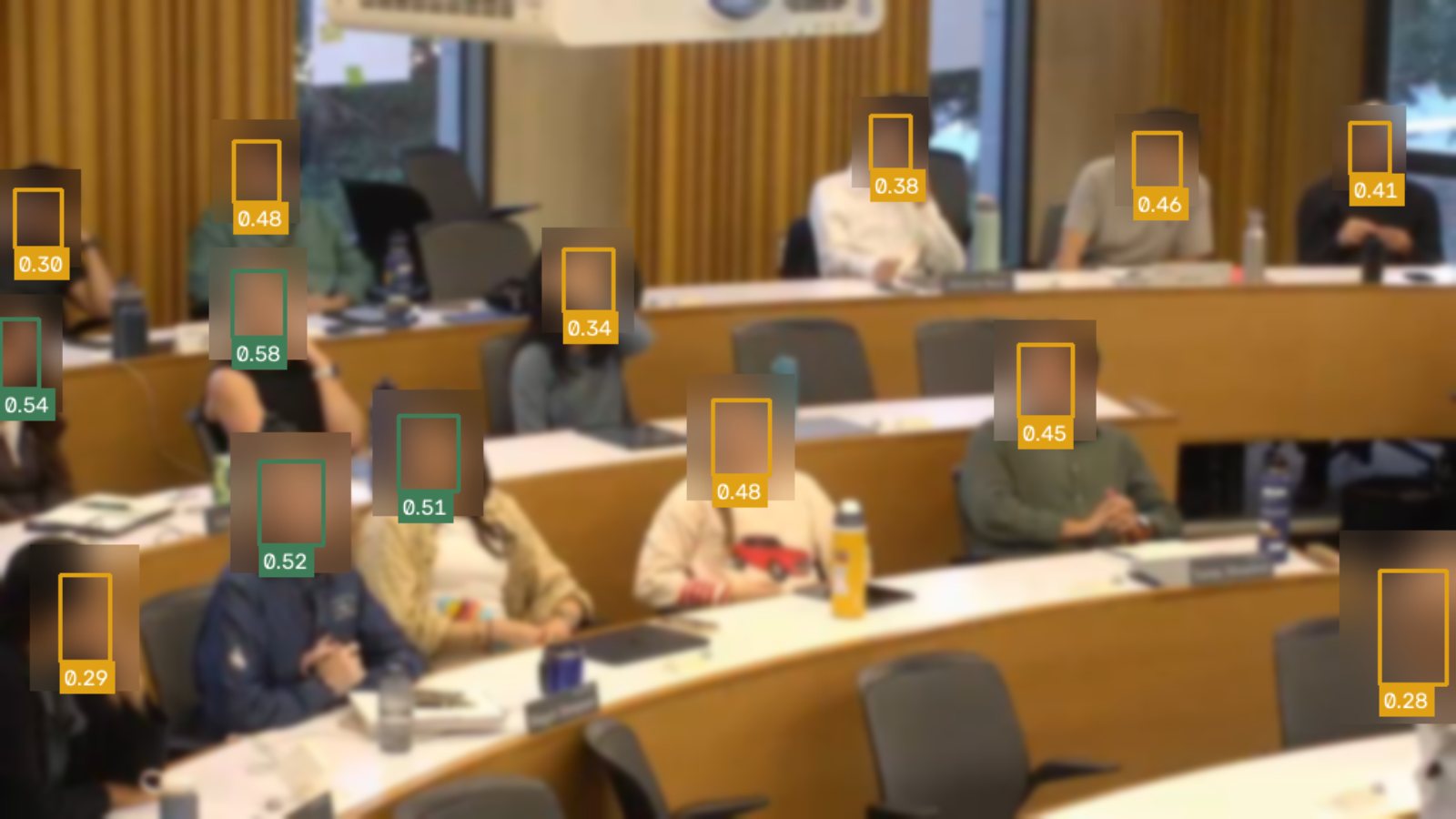}\\[4pt]
\includegraphics[width=0.49\textwidth]{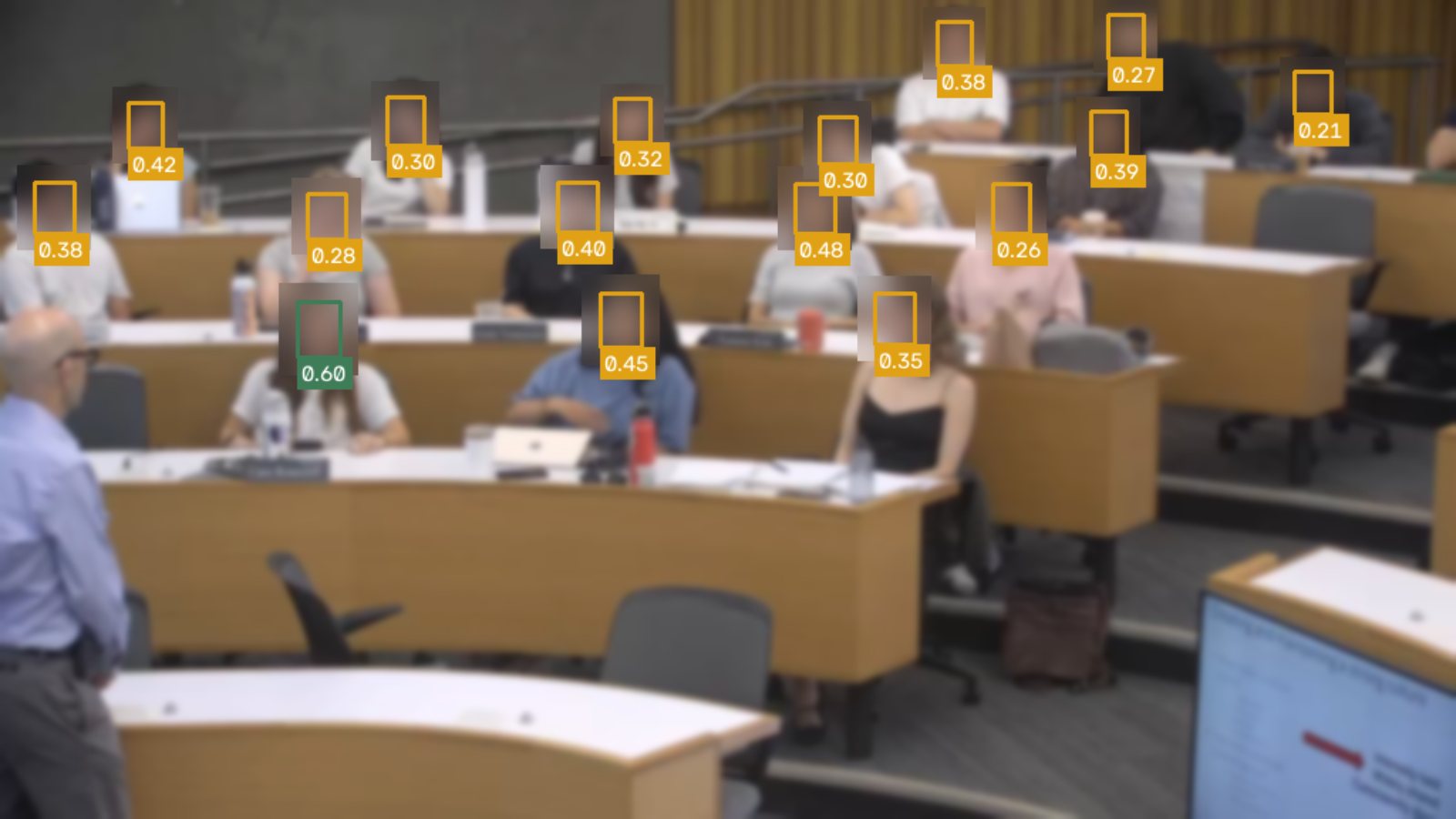}
\caption{Busiest student shots on 24 September, 29 September and 6 October (all 1024$\times$576).}
\end{figure}

\clearpage
\section{Per-student results}
\studenttable

\clearpage
\section{Attendance by student}
\begin{figure}[H]\centering
\includegraphics[width=0.9\textwidth]{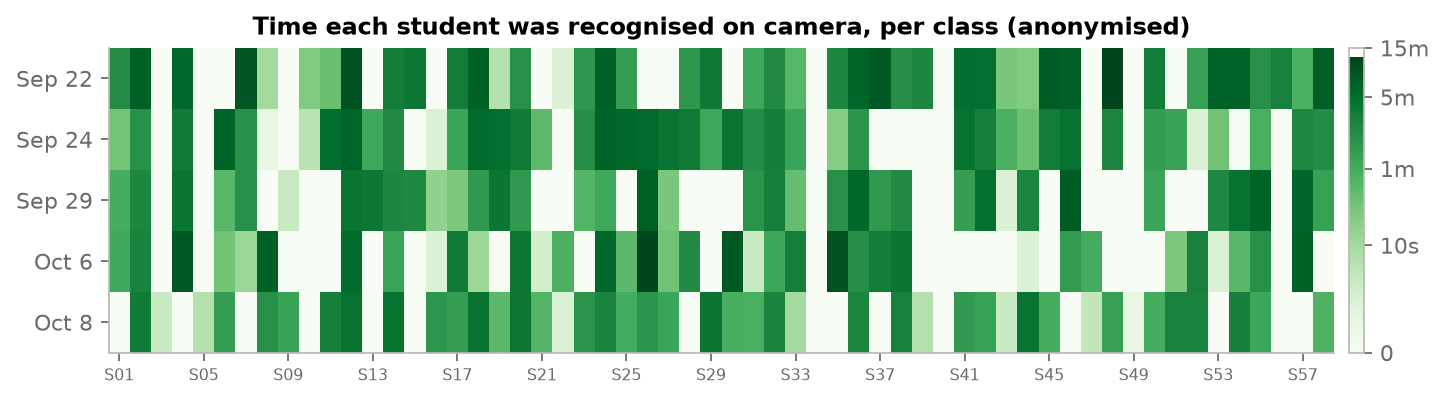}
\caption{Time each student (anonymous id) was recognised on camera, per class.}\label{fig:attname}
\end{figure}

\end{document}